\documentclass{article}

\PassOptionsToPackage{numbers,sort&compress}{natbib} 

\usepackage{soul}

\usepackage{todonotes}

\usepackage{subcaption}  

    \usepackage[preprint]{neurips_2025}

\usepackage[utf8]{inputenc} 
\usepackage[T1]{fontenc}    
\usepackage{hyperref}       
\usepackage{url}            
\usepackage{booktabs}       
\usepackage{amsfonts}       
\usepackage{nicefrac}       
\usepackage{microtype}      
\usepackage{xcolor}         
\usepackage{amsmath} 

\title{Binarized Neural Networks Converge Toward Algorithmic Simplicity: Empirical Support for the Learning-as-Compression Hypothesis}

\author{
Eduardo Y. Sakabe\textsuperscript{1} \\
\texttt{eyujis@gmail.com}
\And
Felipe S. Abrahão\textsuperscript{2,3,4,5} \\
\texttt{felipesabrahao@gmail.com }
\And
Alexandre Simões\textsuperscript{6} \\
\texttt{alexandre.simoes@unesp.br}
\AND
Esther Colombini\textsuperscript{7} \\
\texttt{esther@ic.unicamp.br}
\And
Paula Costa\textsuperscript{1} \\
\texttt{paulad@unicamp.br}
\And
Ricardo Gudwin\textsuperscript{1} \\
\texttt{gudwin@unicamp.br}
\And
Hector Zenil\textsuperscript{3,4,8,9} \\
\texttt{hector.zenil@kcl.ac.uk}
}

\begin{document}

\maketitle

\begin{center}
\textsuperscript{1}School of Electrical and Computer Engineering, University of Campinas (UNICAMP), Brazil \\
\textsuperscript{2}Centre for Logic, Epistemology and the History of Science, University of Campinas (UNICAMP), Brazil \\
\textsuperscript{3}Oxford Immune Algorithmics, Oxford University Innovation \& London Institute for Healthcare Engineering, U.K. \\
\textsuperscript{4}Algorithmic Dynamics Lab, Karolinska Institute \& King’s College London, U.K. \\
\textsuperscript{5}DEXL, National Laboratory for Scientific Computing (LNCC), Brazil \\
\textsuperscript{6}Department of Control and Automation Engineering, São Paulo State University (UNESP), Brazil \\
\textsuperscript{7}Institute of Computing, University of Campinas (UNICAMP), Brazil \\
\textsuperscript{8}Research Departments of Biomedical Computing and Digital Twins, School of Biomedical Engineering and Imaging Sciences \\
\textsuperscript{9}King’s Institute for Artificial Intelligence, King’s College London, U.K.
\end{center}

\begin{abstract}
    Understanding and controlling the informational complexity of neural networks is a central challenge in machine learning, with implications for generalization, optimization, and model capacity. While most approaches rely on entropy-based loss functions and statistical metrics, these measures often fail to capture deeper, causally relevant algorithmic regularities embedded in network structure. We propose a shift toward algorithmic information theory, using Binarized Neural Networks (BNNs) as a first proxy. Grounded in algorithmic probability (AP) and the universal distribution it defines, our approach characterizes learning dynamics through a formal, causally grounded lens. We apply the Block Decomposition Method (BDM)—a scalable approximation of algorithmic complexity based on AP—and demonstrate that it more closely tracks structural changes during training than entropy, consistently exhibiting stronger correlations with training loss across varying model sizes and randomized training runs. These results support the view of training as a process of algorithmic compression, where learning corresponds to the progressive internalization of structured regularities. In doing so, our work offers a principled estimate of learning progression and suggests a framework for complexity-aware learning and regularization, grounded in first principles from information theory, complexity, and computability.

\end{abstract}


\section{Introduction}
\label{sec:introduction}

Understanding the distributional structure of neural network weights from an information-theoretic perspective has driven a range of advances in training efficiency, model compression, and architecture optimization. For example, \cite{susan_dynamic_2014} proposed an entropy-based criterion to dynamically adjust the number of hidden neurons, relating increases in weight entropy to growing representational demands. Similarly, \cite{susan_neural_2019} used the stabilization of weight entropy as a stopping criterion during training. Other approaches incorporate entropy directly into the loss function to regularize complexity and encourage more compact representations \cite{nowlan_simplifying_1992, hinton_keeping_1993, molchanov_variational_2017, wiedemann_entropy-constrained_2019}. Complementarily, post-training compression techniques leverage entropy coding and quantization to reduce model size with minimal impact on accuracy \cite{han_deep_2016, choi_towards_2017, oktay_scalable_2020, wiedemann_compact_2020, wiedemann_deepcabac_2020}.

From the universal (algorithmic) coding theorem (see Section~\ref{sec:preliminaries}) within the context of algorithmic information theory (AIT)~\cite{Downey2010,Chaitin2004,Calude2002,Li1997}, these approaches are often grounded in algorithmic probability~\cite{Zenil2019CausalDeconv,Delahaye2012,zenil_decomposition_2018,HernandezEspinosa2025SuperARC} and universal (Solomonoff) induction~\cite{Li1997,kirchherr1997Miraculous}, such as the minimum description length (MDL) principle~\cite{Li1997, rissanen_stochastic_1986}, 
which posits that the best model is the one that minimizes the total length of two descriptions: the model itself (its parameters or structure) and the data given the model (how well it fits the data). 
Shannon entropy is widely used in this context because it quantifies the expected bit-length required to encode outcomes from a probabilistic source \cite{shannon_mathematical_1948}, directly aligning with MDL. In the case of neural networks, weight entropy serves as a proxy for model complexity, estimating the information needed to represent the network's parameters. Yet, entropy—while widely adopted—captures only statistical variability, overlooking algorithmic and causal structure critical to understanding how neural networks internalize and compress information.
AIT offers an encompassing perspective focused on formal-theoretic measures of complexity (particularly, \textit{algorithmic} (\textit{program-size}) \textit{complexity} and \textit{algorithmic probability}) rather than statistical ones, 
emphasizing the need to capture not just the statistical properties of the data, but also the generative structure (or process) underlying data. 
Universal induction has been regarded as a theoretical solution to Artificial General Intelligence (AGI), positing that the most intelligent systems are those capable of compressing and generalizing via the shortest explanatory programs \cite{HernandezEspinosa2025SuperARC, minsky_limits_2014}. 

We build on this foundation using the Block Decomposition Method (BDM)~\cite{Zenil2019CausalDeconv,zenil_decomposition_2018,Zenil2020cAIDscholarpedia,Zenil2019dAIDcalculusiScience}, a computable and resource effective approximation to the (semicomputable) algorithmic complexity values, to estimate its value in application to the (global) complexity of neural network weights. 
BDM offers a practical and computable approximation that captures both statistical (at global scales) and algorithmic regularities (at local scales), providing a complexity measure that is more granular and more robust to changes in computation models, programming languages, and feature selection characterization of (irreducible/incompressible) information content than entropy is~\cite{zenil_decomposition_2018,LeyvaAcosta2024additively,Zenil2020}.
Such an approach aligns more closely with the theoretical underpinnings of intelligence as algorithmic compression and model synthesis~\cite{lavin_simulation_2022,Zenil2019CausalDeconv,HernandezEspinosa2025SuperARC}. Because the most powerful implementation of BDM operates on binary representations (even when it can also deal with non-binary objects) and can deal with 2D objects such as weight matrices, we binarize network weights and constrain our experiments to Binarized Neural Networks (BNNs)~\cite{hubara_binarized_2016}, which enable its direct application.

In our main experiment, we trained binarized Multilayer Perceptrons (MLPs) \cite{hubara_binarized_2016} with varying hidden layer sizes on MNIST \cite{lecun_gradient-based_1998}, evaluating the correlation between model complexity—measured via BDM and entropy—and training loss across 200 training runs per architecture. Our findings reveal that the Pearson and Spearman correlations between BDM and training loss were consistently higher than those between entropy and loss, suggesting that BDM may serve as a more effective indicator of model complexity and its relationship with training dynamics (see also further discussion in Section~\ref{sec:discussion}). This empirical result supports the broader theoretical claim that training in neural networks can be understood as a process of algorithmic compression \cite{sutskever_simons_2023}---where structure is extracted and encoded in the weights—mirroring the regression/prediction principles proposed e.g. by Solomonoff~\cite{HernandezEspinosa2025SuperARC}. 
These insights highlight the limitations of entropy-based approaches and point to the need for complexity-aware learning principles rooted in algorithmic probability~\cite{Zenil2017aPREentropydeceivinggraphs,Zenil2020}, such as those introduced in~\cite{hernandez-orozco_algorithmic_2021}, which applied BDM to guide learning on non-differentiable spaces.

While related work has primarily examined entropy in terms of layer outputs and mutual information, such as those investigating the bottleneck principle~\cite{tishby_deep_2015, shwartz-ziv_opening_2017, butakov_information_2024}, our approach focuses instead on the complexity of the model itself---specifically the distribution and structure of its weights---rather than the dynamics of input-output mappings in activation values across layers. This distinction matters: weight-based complexity reflects the internal representational capacity and structural organization of the model (which is the case we study in this work); whereas layer output-based measures pertain to how data is processed during inference. 
Our results suggest, and we argue, that understanding and quantifying such an intrinsic complexity of a model is essential not only for interpretability and regularization, but also for advancing theoretical and practical progress toward AGI~\cite{HernandezEspinosa2025SuperARC, minsky_limits_2014}, where learning must reflect causal inference and universal compression rather than statistical fitting alone.

Code and scripts to reproduce our experiments will be made available upon acceptance.

\section{Background}
\label{sec:preliminaries}

\subsection{Basics concepts and results in algorithmic information theory}\label{sectionAIT}

The (unconditional prefix) \emph{algorithmic} (Solomonoff-Kolmogorov-Chaitin) \emph{complexity} of a finite string $ x $, denoted by $ \mathbf{K}(x) $, is the length of the shortest program $x^* \in \mathbf{L_U} $ such that $ \mathbf{K}\left( x \right) =  \left| x^* \right| = \min\left\{ \left| z \right| \; \middle\vert \; \mathbf{U}\left( z \right) = x \right\} $
and $ \mathbf{U}(x^*) = x $,
where $ \mathbf{L_U} $ denote any (prefix-free or self-delimiting) universal programming language running on a machine $ \mathbf{U} $.


Let 
$ \mathbf{P}_{ \mathbf{U} } \left[ x \right] = \sum\limits_{ \mathbf{U}\left( p \right) 
= x } 2^{ - \left| p \right| }  
$
denote the \emph{universal a priori probability} of an arbitrary event $ x $ which can be understood as the probability of randomly generating (by an i.i.d. stochastic process) a prefix-free (or self-delimiting) program that outputs $ x $---in other words,  the probability that event $ x $ occurs resulting from the outcome of at least one of all possible computable generative models, formal theories, computer programs, Turing machines, etc. 

A computably enumerable (c.e.) semimeasure $ \mathbf{m}\left( \cdot \right) $ is said to be \emph{maximal} if, for any other computably enumerable semimeasure $ \mu\left( \cdot \right) $ with domain defined for all possible encoded objects, where $ \sum\limits_{ x \in \left\{ 0 , 1 \right\}^* } \mu\left( x \right) \leq 1 $, there is a constant $ C > 0 $ (which does not depend on $ x $) such that, for every encoded object $ x $,
$ \mathbf{m}\left( x \right) \geq C \, \mu\left( x \right)\text{ .} $

From the \emph{algorithmic coding theorem}~\cite{Chaitin2004,Calude2002,Li1997,Downey2010} (or universal coding theorem) we have that
\begin{equation}\label{equationACT}
	\begin{aligned}
		\mathbf{K}\left( x \right) = 
	- \log\left( 
    \mathbf{P}_{ \mathbf{U} } \left[ x \right]
            \right) \pm \mathbf{O}( 1 )
	= 
	- \log\left( \mathbf{m}\left( x \right) \right) \pm \mathbf{O}( 1 )
	\end{aligned}
\end{equation}
holds,
where
$ p $ denotes a program running on machine $ \mathbf{U} $ such that it returns $ \mathbf{U}\left( p \right) = x $ as output, where $ \left| p \right| $ is the length of the program $ p $.
%
%

We call $ 2^{ - \mathbf{K}\left( x \right) } $ the \emph{algorithmic probability} (AP) of $ x $. 





\subsection{Block Decomposition Method}
\label{subsec:bdm}

The Block Decomposition Method (BDM)~\cite{Zenil2019dAIDcalculusiScience,zenil_decomposition_2018} presents an estimator of algorithmic information redundancies defined by a decomposition of the object into parts for which one already has an algorithmic complexity estimation, obtained by means of, for example, the Coding Theorem Method (CTM)~\cite{Delahaye2012} based on Algorithmic Probability (AP) (see Section~\ref{sectionAIT}) and the related universal distribution~\cite{kirchherr1997Miraculous} which takes into consideration all the statistical and algorithmic regularities and redundancies in data. 
By finding the smallest generating programs (or models), BDM extends the power of CTM by joining these programs together (in a coarse-graining manner) in order to offer a generative computational model of the object, so that one can always achieve tighter lower bounds on AP (or upper bounds on $ \mathbf{K} $) by running CTM further. 
As mentioned in Section~\ref{sectionAIT}, AP gives an agnostic and invariant probability measure for a randomly generated explanation (e.g. a randomly generated computer program) to explain an object~\cite{kirchherr1997Miraculous} or a (-n encodable) set of phenomena so that
it is independent (up to a `small' constant that has been proven to present a stable rate~\cite{Zenil2020,LeyvaAcosta2024additively}) for the chosen computation model, most prominently for low-complexity (or equivalently high algorithmically probable) objects.  
In addition, it demonstrated to be invariant for an arbitrarily chosen programming language, prior probability distribution, and formal theory in the asymptotic limit as the object size increases.

In general case for any encodable multidimensional object~\cite{Ozelim2024AssemblyTheoryReduced}, the BDM of an object $ x $ is  defined by
\begin{equation}\label{eqBDM}
BDM( x , i , m ) = \sum\limits_{ \left( r_j , n_j \right) \in P_i \left( x \right) } \big( \log( n_j ) + K_m( r_j ) \big)
\text{ ,}
\end{equation}
\label{eq:bdm}
\noindent where:
\begin{itemize}
\item the partition (to which one assigns the corresponding index $ i $) is one of the ways to decompose the object $ x $;

\item $ P_i \left( x \right) $ is the set of pairs $ \left( r_j , n_j \right) $ obtained when decomposing the object $ x $ according to a partition $ i $ of contiguous parts $ r_j $, each of which appears $ n_j $ times in such a partition (i.e., $ n_j $ is the multiplicity of $ r_j $), that is, the number of exact repetitions;

\item $  K_m( r_j )  $ is an approximation to $ \mathbf{K}( r_j ) $ and $ m $ is the index of the approximation method employed to calculate $  K_m( r_j )  $.

\end{itemize}

Equation~\eqref{eqBDM} can be expressed for unidimensional objects but also to encodable multidimensional objects in general, such as non-binary strings and $ n $-dimensional objects such as graphs, matrices, images, vectors and tensors~\cite{zenil_decomposition_2018,Zenil2019dAIDcalculusiScience,Zenil2020,Zenil2018a}.
For example, for a bit string $ x $, Equation~\eqref{eqBDM} holds for a partition defined by the sequence of contiguous linear blocks (of length $ \geq 1 $) whose concatenation reconstructs $ x $.


From classical information theory, we have that
\begin{equation}
\mathbf{H}_i\left( X^{ ( i ) } \right)
= 
-
\sum\limits_{ \left( r_j , n_j \right) \in P_i \left( x \right) }
p\left( r_j \right)
\log\left( p\left( r_j \right) \right)
\end{equation}
\label{eq:shannon}

is the \emph{block} (Shannon) entropy $ \mathbf{H}_i $ of an i.i.d. source $ X^{ ( i ) } $ such that $ p\left( r_j \right) \to \frac{ n_j }{N_i } $ as $ \left| x \right| \to \infty $, the random variable $ X^{ ( i ) } $ can assume values in the set $ \left\{ r_1 , \dots , r_j , \dots , r_{ \left| P_i \left( x \right) \right| } \right\} $, and
$ N_i
=
\left(
\sum\limits_{ \left( r_j , n_j \right) \in P_i \left( x \right) } n_j
\right) 
$.

Thus, $ \mathbf{H} $ is a basis for (statistical) compression methods that are subsumed into BDM while for sufficiently large objects both converge to each other.
This is because BDM characterizes the information content of the entire object by adding the estimated (local) complexity given by $ \mathbf{K} $ and the (global) Entropy ($ \mathbf{H} $) values as described in Equation~\eqref{eqBDM}~\cite{Ozelim2024AssemblyTheoryReduced,zenil_decomposition_2018}.

Our results in this paper corroborate these mathematical properties of BDM and entropy, the former expected to perform better for smaller objects while being more sensitive to patterns other than statistical ones.
In Section~\ref{subsec:random_control}, our control experiment evinces the case in which both are indeed expected to converge.

\section{Measuring the Complexity of Binarized Neural Networks}
\label{sec:measuring}
Our primary objective is to investigate whether algorithmic complexity, estimated via the Block Decomposition Method (BDM), serves as a more informative indicator of neural network learning dynamics than entropy. We hypothesize that training a neural network reduces the algorithmic complexity of its weights by aligning them with the structural regularities of the data. In this framing, learning is understood as a form of algorithmic compression: transforming initially random, high-complexity parameters into structured configurations that encode the input-output mappings required by the task.

Accordingly, we expect BDM—which captures local algorithmic regularities beyond statistical variability—to correlate more strongly with training loss than entropy does. While entropy quantifies the expected bit-length under a probabilistic model, it does not account for causal or generative structure within the weight matrix. In contrast, BDM, grounded in algorithmic information theory, approximates algorithmic complexity by detecting repeatable, low-complexity patterns, even in systems that may appear statistically random.

This hypothesis builds on the assumption that effective learning involves the internalization of data structure into the model's parameters in a compact, structured form. We take the training loss as a proxy for this process, assuming that as the loss decreases, the network is increasingly aligned with the regularities in the training data. However, because low loss can also result from memorization rather than generalization, we constrain our analysis to the training regime before the onset of overfitting, as indicated by a plateau in validation loss—where the model is likely compressing the data’s functional structure rather than encoding idiosyncratic details of the training data.

Our approach is consistent with the Minimum Description Length (MDL) principle \citep{rissanen_stochastic_1986}, and related perspectives in deep learning that frame training as a compression process \citep{tishby_deep_2015, schmidhuber_discovering_1997}. By directly comparing BDM and entropy under identical training conditions, we aim to evaluate whether BDM better captures meaningful structural transformations in the model’s parameters during learning.

This view of training as a form of algorithmic compression is supported by recent commentary by Sutskever \citep{sutskever_simons_2023}, who suggests that gradient-based optimization—particularly Stochastic Gradient Descent (SGD)—can be seen as an implicit algorithmic search, uncovering compressed programs within the neural network’s weights. 

\subsection{Computing BDM and Entropy in Binarized Neural Networks}
\label{subsec:computing}

To assess the complexity of a fully connected binarized neural network during training, we compute two measures over its binarized weight matrices: algorithmic complexity using the Block Decomposition Method (BDM), and statistical complexity using entropy. Both measures are derived from a common decomposition of the weights into fixed-size binary submatrices.

\textbf{Weight Extraction and Binarization:}
We extract all weight matrices from the model, excluding batch normalization layers. Each matrix is binarized by applying the sign function, mapping values $>0$ to $1$ and $\leq 0$ to $0$. This produces a set of 2D binary matrices, suitable for pattern-based analysis.

\textbf{Shared Block Decomposition:}
Each binarized matrix is partitioned into non-overlapping $4 \times 4$ blocks. This yields a multiset of binary patterns used as atomic units for both BDM and entropy computation. For each matrix, we count the occurrences of each unique $4 \times 4$ block. These counts define an empirical distribution over the observed patterns.

\textbf{Shannon Entropy:}
The entropy of a matrix is computed using the empirical distribution of $4 \times 4$ patterns as defined in Equation~\ref{eq:shannon}. Here, $p(r_j)$ corresponds to the relative frequency of pattern $r_j$ across all blocks. This entropy captures the statistical variability of local structures in the network's weights.

\textbf{Block Decomposition Method (BDM):}
To estimate algorithmic complexity, we apply BDM as defined in Equation~\ref{eq:bdm}. Each unique $4 \times 4$ block $r_j$ is assigned a complexity value based on the Coding Theorem Method (CTM), and repeated occurrences are penalized logarithmically. This yields a composite complexity score reflecting both diversity and compressibility of local patterns.

\medskip

All computations, including block partitioning, empirical distribution estimation, entropy, and BDM values, were implemented using the \texttt{pybdm} library \citep{talaga_pybdm_2024}.

\section{Experiments}
\label{sec:experiments}
We evaluated the relationship between model complexity and learning dynamics in binarized neural networks trained on MNIST. To assess the robustness of our findings, we also included a control experiment with random binary data and labels, described in Section~\ref{subsec:random_control}.

\subsection{Dataset}

We used the MNIST dataset of handwritten digits \cite{lecun_gradient-based_1998}, a standard image classification benchmark consisting of 28$\times$28 grayscale images of digits from 0 to 9, where the task is to assign each image to its corresponding digit class. We resized each image to $10 \times 10$ pixels to reduce input dimensionality and avoid the domain where BDM approximates entropy. Pixel values were normalized using the dataset mean and standard deviation.

To evaluate generalization and monitor overfitting, we constructed a validation set of 10{,}000 examples, stratified by class, from the original 60{,}000-image training set. The remaining 50{,}000 examples formed a pool from which we generated 200 independent training subsets. Each subset consisted of a stratified random sample of 25{,}000 examples, drawn without replacement, preserving class proportions and ensuring disjointness from the validation set.

The standard MNIST test set (10,000 examples) was held out and used only for reporting final accuracy of the selected models.

\subsection{Model Architecture}
\label{subsec:model}

We used a binarized Multilayer Perceptron (MLP) \cite{hubara_binarized_2016} with two hidden layers, where both weights and activations were binarized using the Straight-Through Estimator (STE) \cite{bengio_estimating_2013}, allowing backpropagation through discrete functions. The model processed resized 10 × 10 pixel MNIST images, with the number of neurons in the two hidden layers denoted as $N_1$ and $N_2$. We applied batch normalization to each hidden layer to stabilize training.

We experimented with different hidden layer configurations, testing $(N_1, N_2)$ pairs of (8, 4), (16, 8), (32, 16), (64, 32), and (128, 64).

\subsection{Training Procedure}
\label{subsec:training}

For each model configuration \((N_1, N_2)\), we trained 200 independent model instances, each on a different stratified subset of the training data. Training employed early stopping with a patience of 5 epochs based on validation loss: if the validation loss did not improve for 5 consecutive epochs, training was halted, and the model with the lowest validation loss was selected.

We optimized the cross-entropy loss using the Adam optimizer with a learning rate of \(1 \times 10^{-3}\) and a mini-batch size of 128.

To enable post hoc analysis of model complexity and entropy, we saved the model weights after every training step—that is, after each backpropagation update. BDM and entropy were computed at this resolution and later averaged per epoch to align with the training and validation loss: the former was averaged over batches, and the latter was computed once per epoch.

All experiments were run on a MacBook Pro with an M4 Max chip using PyTorch with Metal acceleration. Training 200 models for the largest architecture took about 5 hours, and BDM/entropy computations required an additional 2.5 hours.

\subsection{Evaluation}
\label{subsec:evaluation}

We assessed the relationship between model complexity and learning dynamics by analyzing the correlations between mean training loss and mean complexity metrics (BDM and entropy) computed per epoch throughout training. Our evaluation followed three main steps:

\textbf{Metric Normalization}: 
To ensure comparability and reduce noise, we applied a normalization pipeline to the per-epoch series of training loss, BDM, and entropy values. We excluded the final five epochs prior to early stopping to avoid the overfitting regime. We applied log transformation, Gaussian smoothing, and MinMax scaling to the remaining values.
    
\textbf{Correlation Analysis}: 
Using the normalized values, we computed Pearson and Spearman correlation coefficients between training loss and each complexity metric. These correlations quantified both linear (Pearson) and monotonic (Spearman) relationships, providing insight into how closely each metric tracked the progression of learning.

\textbf{Bootstrap Confidence Intervals}: 
To assess statistical reliability, we estimated 95\% confidence intervals via bootstrap resampling over the 200 independently trained models for each architecture. This provided robust estimates of variability for all reported correlations.

\subsection{Results}
\label{subsec:results}

The results of our experiments are summarized in Table~\ref{tab:results}, which presents 95\% confidence intervals for Pearson and Spearman correlations between model complexity---measured using BDM and entropy---and the mean training loss across different model configurations. We also reported the final test accuracies of the selected models based on the lowest validation loss. The configurations of hidden layers were denoted by $N_1, N_2$, where $N_1$ is the number of neurons in the first hidden layer and $N_2$ in the second.

\begin{table}[htbp]
\caption{Correlation [95\% CI] between training loss and complexity metrics across model sizes, with final test accuracy reported as mean ± standard deviation. $r$ and $\rho$ denote Pearson and Spearman correlations, respectively. Bold values indicate the higher correlation in each pair. BDM consistently outperforms entropy across all models.}
\label{tab:results}
\centering
\small
\begin{tabular}{cccccc}
\toprule
$N_1$, $N_2$ & BDM $r$ & Entropy $r$ & BDM $\rho$ & Entropy $\rho$ & Accuracy (\%) \\
\midrule
8, 4       & \textbf{[0.72, 0.81]} & [0.45, 0.53] & \textbf{[0.59, 0.70]} & [0.50, 0.62] & 51.5 ± 5.1 \\
16, 8      & \textbf{[0.85, 0.89]} & [0.55, 0.63] & \textbf{[0.74, 0.81]} & [0.55, 0.66] & 67.8 ± 2.3 \\
32, 16     & \textbf{[0.90, 0.92]} & [0.69, 0.74] & \textbf{[0.73, 0.80]} & [0.61, 0.70] & 79.8 ± 1.1 \\
64, 32     & \textbf{[0.91, 0.92]} & [0.82, 0.83] & \textbf{[0.84, 0.88]} & [0.79, 0.84] & 85.9 ± 0.5 \\
128, 64    & \textbf{[0.91, 0.91]} & [0.85, 0.87] & \textbf{[0.90, 0.93]} & [0.88, 0.91] & 89.0 ± 0.3 \\
\bottomrule
\end{tabular}
\end{table}

Correlations were consistently stronger for BDM than entropy, particularly in smaller models, where the influence of algorithmic structure was more pronounced.

To ensure the reliability of the correlation estimates, we excluded runs that terminated before 10 epochs, allowing for a consistent 5-epoch window preceding early stopping. Consequently, the number of runs included in the bootstrap analysis was 175 for the (8, 4) model and 195 for the (16, 8) model, while all other configurations retained the full set of 200 runs. These sample sizes remained sufficient for robust statistical inference.

\begin{figure}[t]
    \centering

    \begin{subfigure}[t]{0.48\linewidth}
        \centering
        \includegraphics[width=\linewidth]{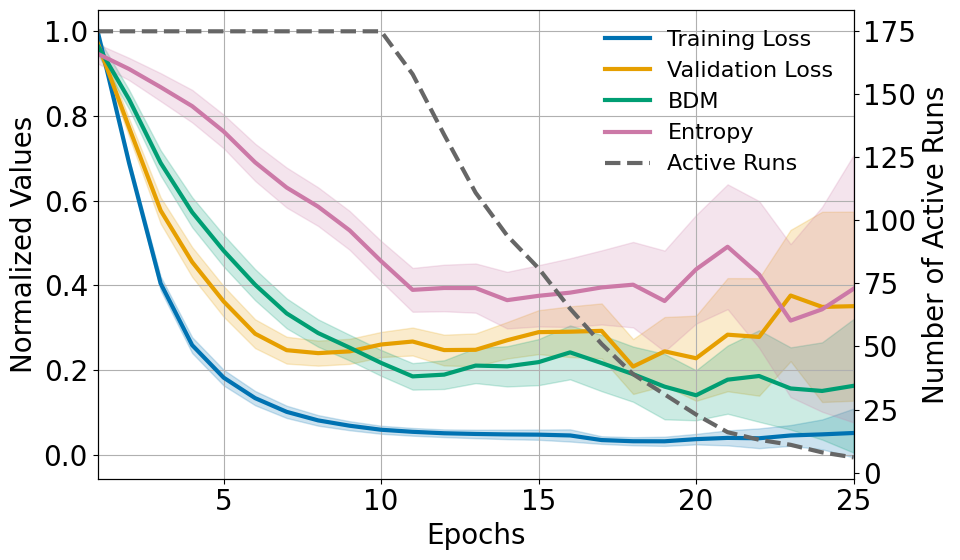}
        \caption{\scriptsize Architecture (8, 4)}
    \end{subfigure}
    \hspace{0.02\linewidth}  
    \begin{subfigure}[t]{0.48\linewidth}
        \centering
        \includegraphics[width=\linewidth]{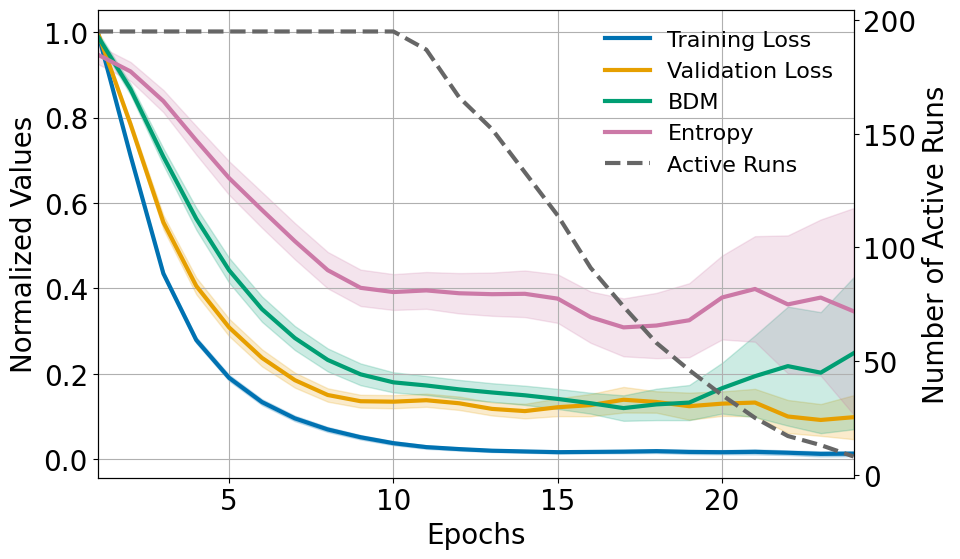}
        \caption{\scriptsize Architecture (16, 8)}
    \end{subfigure}

    \vspace{2ex}

    \begin{subfigure}[t]{0.48\linewidth}
        \centering
        \includegraphics[width=\linewidth]{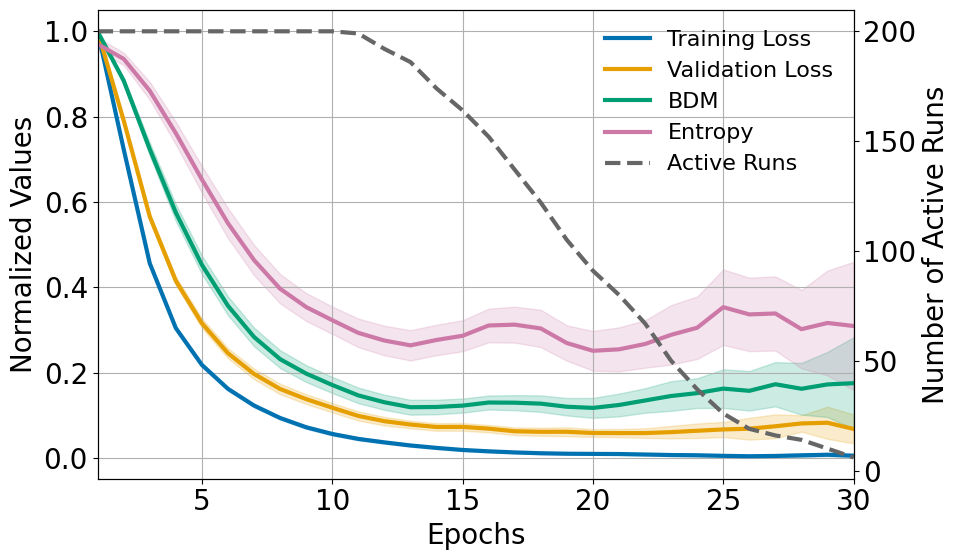}
        \caption{\scriptsize Architecture (32, 16)}
    \end{subfigure}
    \hspace{0.02\linewidth}
    \begin{subfigure}[t]{0.48\linewidth}
        \centering
        \includegraphics[width=\linewidth]{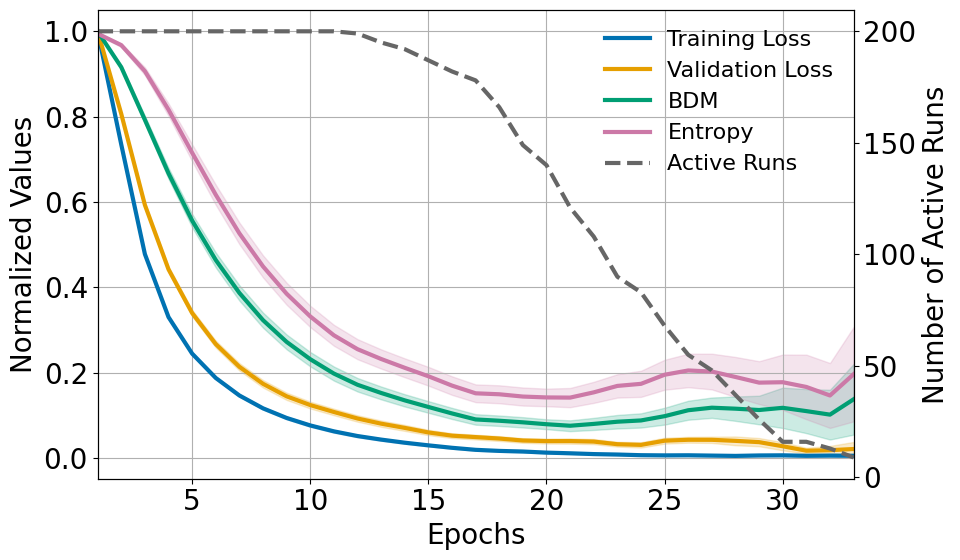}
        \caption{\scriptsize Architecture (64, 32)}
    \end{subfigure}

    \vspace{2ex}
    \hspace*{-0.06\linewidth} 
    \begin{subfigure}[t]{0.48\linewidth}
        \centering
        \includegraphics[width=\linewidth]{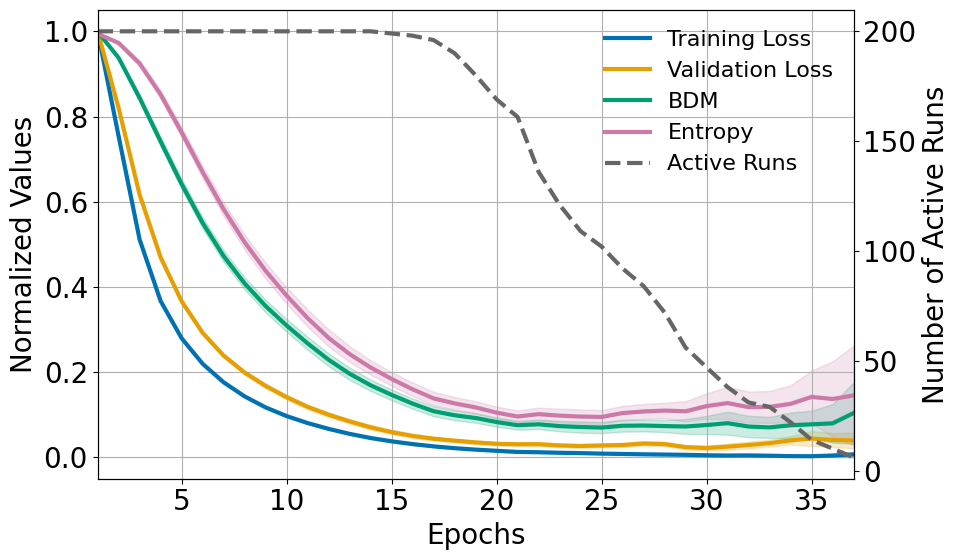}
        \caption{\scriptsize Architecture (128, 64)}
    \end{subfigure}
    \hspace{0.02\linewidth}
    \begin{subfigure}[t]{0.43\linewidth}
        \centering
        \includegraphics[width=\linewidth]{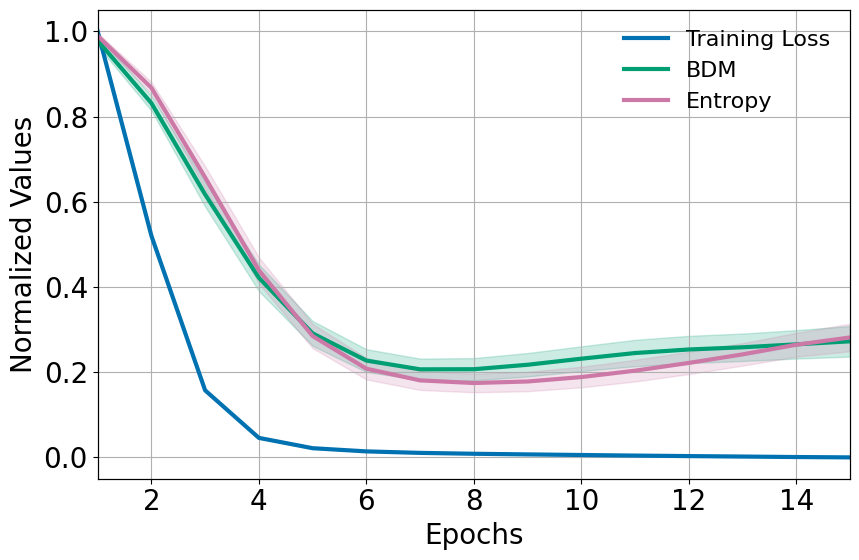}
        \caption{\scriptsize Architecture (32, 16) trained on \textbf{random} inputs}
    \end{subfigure}

    \caption{
    Evolution of training loss, validation loss, BDM, and entropy across epochs for each model architecture. Subplots (a)–(e) correspond to increasing model sizes trained on MNIST, while (f) shows the control condition using random inputs with the (32, 16) architecture. All metrics are normalized to enable direct comparison across runs. Shaded regions represent 95\% confidence intervals, computed only for epochs with at least five active runs. The dashed gray line indicates the number of active runs per epoch, reflecting early stopping behavior. Across all architectures, BDM more closely follows the trajectory of training loss than entropy, particularly during early and mid-training. This alignment suggests that BDM is more sensitive to the structural changes induced by learning. BDM also exhibits lower variance across runs, providing more stable complexity estimates throughout training. In the random-data condition (f) from our control experiment (see Sections~\ref{subsec:random_control} and~\ref{sec:discussion}), where no meaningful structure is present, the distinction between BDM and entropy largely vanishes—reinforcing the interpretation that BDM’s advantage depends on its sensitivity to underlying algorithmic regularities.
    }
    \label{fig:training_dynamics}
\end{figure}

To complement the aggregate correlation results presented in Table \ref{tab:results}, we visualized the evolution of BDM and entropy values alongside training and validation loss across training epochs for each model architecture in Figure \ref{fig:training_dynamics}a-e.

\subsection{Control Experiment with Random Data}
\label{subsec:random_control}

To assess whether the observed correlations between complexity and training loss reflected meaningful structure in the data, we conducted a control experiment using a synthetic dataset consisting of $10 \times 10$ random float inputs (sampled uniformly from $[0, 1)$) and randomly assigned class labels. This design ensured that any learning reflected memorization rather than compression of structured information.

We used the same (32, 16) architecture and training procedure as in the corresponding MNIST experiment, with 200 class-balanced subsets to control for label imbalance effects. We chose this configuration because it represented a mid-sized model where BDM consistently outperformed entropy. Unlike the main experiments, models were trained for a fixed duration of 15 epochs without early stopping, as the absence of shared structure between training and validation sets rendered validation loss uninformative for model selection or generalization; accordingly, it was not computed during training.

We computed correlations between training loss and both complexity measures across all runs to evaluate their behavior in the absence of learnable structure. Despite the lack of meaningful patterns, nonzero correlations may still emerge due to memorization. In this random-data setting, the 95\% confidence intervals were: BDM — Pearson [0.72, 0.77], Spearman [0.43, 0.54]; Entropy — Pearson [0.76, 0.80], Spearman [0.46, 0.55]. The corresponding training dynamics are illustrated in Figure~\ref{fig:training_dynamics}f. We discuss these results in the following section.

\section{Discussion}
\label{sec:discussion}

We first verified that all models performed significantly above chance on MNIST. While a random classifier achieves roughly 10\% accuracy, even the smallest architecture exceeded 50\%, indicating that the networks have learned meaningful input-output mappings from the data.

The correlation results reveal strong positive relationships between model complexity and training loss for both BDM and entropy. These findings indicate that as models reduce error over time, their structural and statistical complexity also decrease. Across all configurations, BDM exhibits higher Pearson and Spearman correlation coefficients than entropy, suggesting that BDM is more sensitive to changes in the model throughout training, particularly in smaller architectures where algorithmic regularities are more distinct.

The higher Pearson correlations imply that BDM tracks the magnitude of changes in training loss more closely. Concurrently, the stronger Spearman correlations indicate that BDM better preserves the relative ordering of complexity over training epochs. Together, these results suggest that BDM provides a richer signal of learning progression than entropy, likely due to its foundation in algorithmic information theory, which captures more than just statistical regularities (see Section~\ref{sec:preliminaries}).

However, this advantage diminishes as model size increases, since BDM relies on evaluating fixed-size \(4 \times 4\) binary blocks via the CTM. As the size of the network grows, the decomposition process leads to increasing redundancy and a heavier influence of the multiplicity term \(\log_2 n_i\) in Equation~\ref{eq:bdm}. This results in a loss of granularity and a convergence of BDM toward entropy-like behavior, reducing its ability to discriminate structural complexity. Thus, in larger models, BDM transitions from an algorithmic to a more statistical measure (see Section~\ref{sec:preliminaries} and a discussion on limitations in Section~\ref{subsec:limitations}).

The training dynamics shown in Figure~\ref{fig:training_dynamics}a-e further support these findings. Across all architectures, BDM exhibits a trajectory that more closely follows the evolution of training loss compared to entropy. This temporal alignment reinforces the view that BDM is more responsive to the structural transformations that occur as the model learns. Moreover, although confidence intervals naturally widen toward later epochs due to early stopping and reduced sample sizes, entropy displays greater variability across runs at each epoch. This difference in variance suggests that BDM not only correlates more strongly with training loss but also produces more stable complexity estimates during training.

In the control experiment with random data, the model successfully minimized training loss over 15 epochs, demonstrating its capacity to memorize arbitrary inputs in the absence of learnable structure. Compared to the MNIST setting, BDM correlations with training loss were reduced, while entropy correlations remained relatively stable and slightly higher. This contrast reinforces the interpretation that BDM is specifically sensitive to compressible structure, whereas entropy continues to reflect general memorization.

In this unstructured condition, entropy exhibited slightly higher correlations than BDM across both Pearson and Spearman metrics. This reversal relative to the MNIST results suggests that, in the absence of algorithmic regularities, BDM converges toward statistical behavior and loses its advantage. As shown in Figure~\ref{fig:training_dynamics}f, the trajectories of BDM and entropy are closely aligned, reflecting the lack of structural signals and supporting the view that BDM’s value emerges only in the presence of causal or generative patterns in the data.

\subsection{Limitations}
\label{subsec:limitations}

While algorithmic information theory provides a more principled and causally grounded foundation for characterizing neural network complexity, our approach has several limitations that constrain its broader applicability.

First, algorithmic complexity is inherently non-continuous and non-differentiable. This prevents its direct integration into gradient-based optimization algorithms such as backpropagation, making it unsuitable as a training-time regularizer in its current form. Second, the current implementation of the Block Decomposition Method (BDM) requires binary inputs in two-dimensional structures. Consequently, our analysis is restricted to Binarized Neural Networks (BNNs), which, while useful as a simplified model class, are less commonly used than full-precision architectures in practical applications. Third, for larger and more complex objects, BDM tends to converge toward entropy, as discussed earlier. This convergence reduces its sensitivity to deeper algorithmic structure in high-dimensional settings and limits its advantage over entropy-based measures in such cases.

Taken together, these limitations currently preclude the direct application of our method to large-scale, non-binarized models. Overcoming these constraints—either by developing differentiable approximations of algorithmic complexity or by extending BDM to richer data representations—remains an important direction for future work.

\section{Conclusion}
\label{sec:conclusion}

This work presents a principled investigation of neural network training through the lens of algorithmic information theory. By applying the Block Decomposition Method (BDM) to Binarized Neural Networks (BNNs), we demonstrated that algorithmic complexity offers a more sensitive and stable indicator of training dynamics than traditional entropy.

Empirical results across multiple architectures show that BDM correlates more strongly with training loss than entropy, particularly in smaller models, where algorithmic regularities are more pronounced. These findings offer direct empirical support for the view that training operates as a process of algorithmic compression, transforming random initializations into structured, compressible configurations that reflect the underlying data-generating process. Control experiments with random-input data further reinforce this interpretation: in the absence of meaningful structure, the advantage of BDM disappears, and its behavior converges toward that of entropy. These results confirm that BDM captures structural features intrinsic to learning, beyond distributional statistics.

Taken together, our results highlight the potential of algorithmic complexity measures to enrich our understanding of neural network behavior. They open new directions for future work, including the development of complexity-aware training regimes, regularization strategies based on algorithmic information theory, and the design of learning systems that exploit causal-compressibility as a guiding principle.  This perspective is especially relevant in the context of emerging architectures characterized by localized computation—such as sparse neural networks \cite{frankle_lottery_2019}, transformers \cite{vaswani_attention_2017}, Mixture-of-Experts (MoE) models \cite{shazeer_outrageously_2017}, and Kolmogorov–Arnold Networks (KANs) \cite{liu_kan_2025}. In these systems, BDM may have an even greater advantage, as it is particularly well-suited for characterizing modular structures \cite{hernandez-orozco_algorithmically_2018}.

Our work, alongside that of \cite{hernandez-orozco_algorithmic_2021}, begins to address a longstanding challenge: integrating algorithmic complexity and algorithmic probability—long proposed as a theoretical solution to AI through universal induction—into practical machine learning. Despite their foundational role in the theoretical foundations of artificial intelligence, these concepts have remained largely disconnected from neural network training. By operationalizing algorithmic complexity via BDM in binarized architectures, we take a step toward bridging this gap—replacing statistical proxies like entropy with causally grounded, algorithmic measures. In doing so, we contribute to realizing algorithmic theories of learning in practice, bringing foundational principles of AI closer to their application in modern machine learning.

\section{Acknowledgements}
This project was supported by the Ministry of Science, Technology, and Innovation of Brazil, with resources granted by the Federal Law 8.248 of October 23, 1991, under the PPI-Softex. The project was coordinated by Softex and published as Intelligent agents for mobile platforms based on Cognitive Architecture technology [01245.003479/2024-10]. This study was partially funded by the Coordenação de Aperfeiçoamento de Pessoal de Nivel Superior – Brasil (CAPES) – Finance Code 001.
Felipe S. Abrah\~{a}o acknowledges support from the São Paulo Research Foundation (FAPESP), grants $2021$/$14501$-$8$.

\small

\end{document}